\documentclass{article}

\usepackage[nonatbib, final]{neurips_2019}

\usepackage[utf8]{inputenc} 
\usepackage[T1]{fontenc}    
\usepackage{hyperref}       
\usepackage{url}            
\usepackage{booktabs}       
\usepackage{amsfonts}       
\usepackage{nicefrac}       
\usepackage{microtype}      
\usepackage{soulutf8}
\usepackage{color}
\usepackage{bm}
\usepackage[caption=false]{subfig}
\usepackage[export]{adjustbox}
\usepackage{tabularx}
\usepackage{epstopdf}
\usepackage{amsmath}

\title{Efficient Neural Task Adaptation by Maximum Entropy Initialization}
\author{%
 Farshid Varno \\ 
  Dalhousie University\\
  Imagia Cybernetics \\
  \texttt{f.varno@dal.ca}
     \And
  Behrouz Haji Soleimani \\
  Dalhousie University\\
  \textit{behrouz.hajisoleimani@dal.ca}\\
  \And
  Marzie Saghayi \\
  Dalhousie University\\
  \textit{m.saghayi@dal.ca}\\
  \And
  Lisa Di Jorio\\
  Imagia Cybernetics \\
  \texttt{lisa@imagia.com} \\
  \And
  Stan Matwin \\
  Dalhousie University\\
  Polish Academy of Sciences\\
  \textit{stan@cs.dal.ca}\\
}

\begin{document}
\newcommand{\Lagr}{\mathcal{L}}

\maketitle

\begin{abstract}
Transferring knowledge from one neural network to another has been shown to be helpful for learning tasks with few training examples. Prevailing fine-tuning methods could potentially contaminate pre-trained features by comparably high energy random noise. This noise is mainly delivered from a careless replacement of task-specific parameters. We analyze theoretically such knowledge contamination for classification tasks and propose a practical and easy to apply method to trap and minimize the contaminant. In our approach, the entropy of the output estimates gets maximized initially and the first back-propagated error is stalled at the output of the last layer. Our proposed method not only outperforms the traditional fine-tuning, but also significantly speeds up the convergence of the learner. It is robust to randomness and independent of the choice of architecture. Overall, our experiments show that the power of transfer learning has been substantially underestimated so far.
\end{abstract}
\section{Introduction}
Artificial Neural Networks (ANNs) have shown a great capacity in learning complicated tasks and have become the first contender to solve many problems in the machine learning community. However, 
a large training dataset is a key prerequisite for these networks to achieve good performances. This limitation has opened a new chapter in neural network research, which attempts to make learning possible with limited amounts of data. 
So far, one of the most widely used techniques to cope with such hindrance is the initialization of parameters based on the prior knowledge obtained from already trained models. 

To adapt a pre-trained model to a new task, usually task-specific, extraneous and random parameters are transplanted to a meaningful set of representations, resulting in a heterogeneous model \cite{girshick2014rich, sharif2014cnn, agrawal2014analyzing, li2018learning}. Training these unassociated modules together could contaminate the genuinely learned representations and significantly degrade the maximum transferable knowledge. Current fine-tuning techniques slow down the training process to compensate for this knowledge leak~\cite{li2018learning} which undermines fast convergence of a model that suffers from data-shortage.

We address this dilemma by introducing Efficient Neural Task Adaptation by Maximum Entropy initialization (ENTAME). ENTAME significantly decreases the initial noise that is back-propagated from randomly-initialized parameters toward layers that contain the transferred knowledge. Our experiments show that models trained by the proposed method learn substantially faster and almost always outperform those using traditional fine-tuning or even more complicated tricks like warm up~\cite{li2018learning}. ENTAME is easy to implement and can be beneficially applied to any pre-trained neural network that estimates output probabilities using softmax logits.

In this paper, the link between parameter initialization and the energy of back-propagated error is analyzed theoretically for neural networks using Cross Entropy (CE) loss with softmax logits. 
We derive the optimal parameter initialization for neural networks being fine-tuned on pre-trained models for classification and show that such optimal initial loss leads to a significant acceleration in adapting a pre-trained neural network to a new task.
Our method is independent of the choice of architecture and could be applied to transfer knowledge within any domain. To the best of our knowledge, this is the first work that investigates initialization for task adaptation. Based on our results, ENTAME has a significant practical impact on convergence.
Our main contributions in this paper are as follows:

\begin{itemize}
    \item We decompose the total energy of back-propagated error $\phi$ into three components: the energy of the labels, the energy of the output estimates, and an accuracy-related term. We then find bounds for each component.
    \item We provide a simple yet powerful solution that maximizes the initial entropy of the output estimates and prevents knowledge contamination in the transfer learning setting. The concepts that we introduce are universal and independent of the choice of architecture.
    \item We inspect the practical difference between typical fine-tuning and training using warm-up with our proposed method and empirically show that our method significantly accelerates the convergence of pre-trained models.
\end{itemize}

\section{Related works}
Previous studies on parameters initialization of ANNs \cite{glorot2010understanding, he2015delving, arpit2019benefits, klambauer2017selfnormalizing} focus on preserving the variance (or other statistics) of the flowing data along the depth. This stabilizes the model and makes training deeper networks possible. Arpit and Bengio~\cite{arpit2019benefits} recently showed that the initialization introduced in~\cite{he2015delving} is the optimal one for a ReLU network trained from scratch.~\cite{arpit2019benefits} recommended to use the fan-out mode which preserves the variance of the back-propagated error along the depth. He et al.~\cite{he2015delving}, inconsistently exempted the last layer of the models used for their experiments from the distribution for weights that they have recommended. This layer's distribution is stated to be found experimentally and no justification has been provided for its outcome. Such a strategy could be traced down to the earlier practices in constructing deep neural networks~\cite{simonyan2014deep}. Instead of speculation, we investigate the effect of the initialization of the last layer on the training procedure. However, we narrow down our focus to the optimal initialization for adapting a pre-trained model to a new task. 

Recent studies on transfer learning use variance preserving initialization techniques for fine-tuning \cite{li2018learning, shermin2018transfer}. However, we show that using such techniques, initially contaminates the transferred knowledge, resulting in unguided modification of valuable transferred features. 

Careful initialization is also an inevitable part of self-normalized neural networks introduced in~\cite{klambauer2017selfnormalizing}. These networks use Scaled Exponential Units (SELUs) as their activation function. We restrain our attention to ReLU networks while the concepts that we introduce are independent of the architecture and initialization and could be universally applied to boost the performance of any Feed-forward Neural Network (FNN).

In feature extraction \cite{donahue2013decaf, azizpour2016factors}, the pre-trained features are only used in inference mode and corresponding parameters remain intact during training. This protects the learned representations from undesired contamination but also prevents the required new task-specific features to be learned.

Fine-tuning~\cite{girshick2014rich} lets the pre-trained features and augmented parameters learn the target task together. Fine-tuning usually performs better than feature extraction and training from scratch with random initialization~\cite{girshick2014rich}. However, the pre-trained features are substantially contaminated
due to noise flowing from random layers to the loss and from there, back-propagated toward the features. This paper tackles this problem by eliminating this noise within the augmented parameters. 

\section{Background}
\subsection{Flow of data and error}
An FNN is usually built by stacking up a number of layers on top of each other. The input of a layer can be composed of any combination of the previous layers' outputs. Let the input and output of the $l$-th layer of an FNN having $L$ layers be $\bm{X}^{l}$ and $\bm{A}^{l}$ respectively. They are related to each other through functions $g(.)$ and $h(.)$ as

\begin{equation}
    \bm{X}^{l} =  g^{l} \left( \bm{A}^{1}, \bm{A}^{2}, \ldots \bm{A}^{l-1} \right) \ ; \ \bm{A}^{l} =  h^{l} \left( \bm{X}^{l}, \bm{W}^{l}, \bm{b}^{l}\right) \ ,
\end{equation}
where $\bm{W}^{l} \in \mathbb{\mathop{R}}^{U\times V}$ and $\bm{b}^{l} \in \mathbb{\mathop{R}}^{1 \times V}$ are weights and biases of the $l$-th layer and $V$ is the number of columns in $\bm{X}^{l}$.

If the target task is classification with $C$ classes, then the last layer is usually a fully connected one with $\bm{A}^{L} \in \mathbb{\mathop{R}}^{N\times C}$, where $N$ is the number of examples passing through the network, known as \textit{batch size}. Specifically for this layer

\begin{equation}
    \bm{A}^{L} = \bm{X}^{L}  {\bm{W}^{L}}^{T} + \bm{\overrightarrow{1}}\bm{b}^{L},
    \label{eq:A_L}
\end{equation}
where $\bm{\overrightarrow{1}}$ is a column vector of ones. 

Since many formulas in this paper are batch independent and could be easily broadcast, we separately use lower-case bold-face letters of corresponding introduced matrices to indicate a single example in the batch ($N=1$).
Therefore, Equation~\ref{eq:A_L} could be rewritten for a single example as 
\begin{equation}
    \bm{a^L} = \bm{x^L} {\bm{W}^{L}}^{T} + \bm{b}^{L}.
\end{equation}

The posterior of the last layer's neurons, corresponding to each class, is usually estimated using a softmax normalizer, defined as

\begin{equation}
    {\hat{y}}_{j} = \frac{e^{{a}^{L}_{j}}}{\sum\limits_{i=1}^{C} e^{{a}^{L}_{i}}} \ ,
    \label{eq:sm}
\end{equation}
where $a^{L}_{j}$ represents the $j$-th element of $\bm{a}^{L}$.

Cross entropy is the most commonly used loss function for classification tasks and is equal to the Kullback-Leibler divergence between the labels and the estimates, $\Lagr = - D_{\textit{KL}}(\bm{Y},\bm{\hat{Y}})$.
To train the network using back-propagation~\cite{hornik1989multilayer}, gradients of the loss with respect to each parameter are calculated. To make this easier using the chain rule, first, the gradients are computed with respect to the output of each layer as in 
\begin{equation}
\bm{\delta}^{l} = \nabla_{\bm{a}^l} \Lagr = \left[\frac{\partial \Lagr}{\partial {a}^{l}_1}, \frac{\partial \Lagr}{\partial {a}^{l}_2}, \ldots \right], 
\label{eq:delta_define}
\end{equation}
and from there the desired gradients are calculated;
\begin{equation}
    \frac{\partial \Lagr}{\partial \bm{W}^{l}_j} = \bm{\delta}^{l}_j \frac{\partial h^{l}}{\partial \bm{W}^{l}_j}\  ; \ \frac{\partial \Lagr}{\partial b^{l}_j} = \bm{\delta}^{l}_j \frac{\partial h^{l}}{\partial b^{l}_j} \ , 
    \label{eq:pat_L2w_j}
\end{equation}
where $\bm{W}^{l}_j$ is the $j$-th row of $\bm{W}^{l}$. 

The gradients of the CE loss with respect to the output of the $j$-th neuron of the deepest layer are equal to

\begin{equation}
{\delta}^{L}_j = \frac{\partial \Lagr}{\partial {a}^{L}_j} = \frac{1}{N} \left( {\hat{y}}_j - {y}_j \right)
\label{eq:delta_last}
\end{equation}
and since the last layer is fully connected, the back-propagated error to the previous layer is also easily calculated using the chain rule:
\begin{equation}
    {\delta}^{L-1}_k = \sum\limits_{j=1}^{C} {\delta}^{L}_j {W}^{L}_{j,k},
    \label{eq:delta2delta}
\end{equation}
where ${W}^{L}_{j,k}$ is the $k$-th element of $\bm{W}^{L}_{j}$.
Finally, the weights and biases are updated using gradient descend.

Specifically for the last layer, the gradients of loss with respect to the rows of the weight matrix are
\begin{equation}
    \nabla_{\bm{W}^{L}_j} \Lagr = N \mathop{\mathbb{E}_N} \left[{{\delta}^{L}_j}\bm{x}^{L}\right],
    \label{eq:part_l_wj_expansion}
\end{equation}
where $\mathop{\mathbb{E}_N}$ is the expectation over the examples in the batch. Likewise the derivative of the loss with respect to a single bias in the last layer is
\begin{equation}
    \frac{\partial \Lagr}{\partial {b}^{L}_j} = N \mathop{\mathbb{E}_N} \left[{\delta}^{L}_j\right]. 
    \label{eq:part_l_bj_expansion}
\end{equation}

\subsection{Initialization}
During the back-propagation algorithm~\cite{hornik1989multilayer}, each data entry is passed twice through each layer's weights except the layers fed directly by the raw input. The magnitude of weights in a layer may get affected by the energy of the input visited by the layer, and the error back-propagated up to its output. Mathematically, it means that in Equations \ref{eq:pat_L2w_j}, a term of $\bm{x}^{l}$ usually appears in the derivative of $h^{l}$ with respect to the weights of the $l$-th layer. This is already shown for the last layer in Equation~\ref{eq:part_l_wj_expansion}. Weights are distinguished from biases and called 
such since they involve multiplication. This operation can rapidly increase/decrease the energy of its result, compared to the operands. This intensified/lessened energy of the output may increase/decrease the energy of the weights themselves through the gradient updates as discussed. This loop can lead to numerical problems known as exploding/vanishing gradients. One way of facing these problems is to initialize the weights and biases such that the energy of the flowing data/error is preserved. Currently, energy preserving initialization~\cite{he2015delving} is known to be the optimal solution for training ReLU networks~\cite{arpit2019benefits}.

\subsection{Problem Statement}
\begin{figure}[t]
\centering
  \subfloat[MNIST\label{fig:xl_var_mnist}]{
      \includegraphics[width=0.40\textwidth]{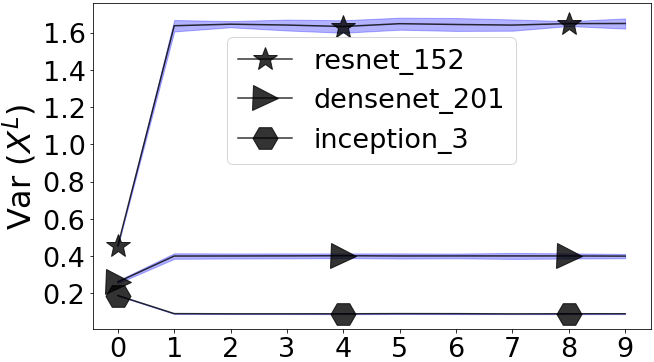}}
  \subfloat[CIFAR100\label{fig:xl_var_cifar100} ]{
      \includegraphics[width=0.41\textwidth]{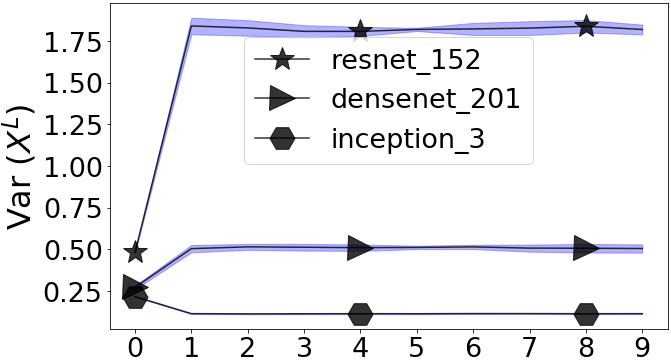}}
\caption{\label{fig:xl_vars} Sudden initial change in variance of $\bm{X}^{L}$ with initial learning rate equal to 0.0001, fine-tuned on (a) MNIST and (b) CIFAR100 datasets. The horizontal axes show the training steps. In each model the augmented parameters are  initialized based on preserving the variance of gradient recommended in~\cite{he2015delving}. The color shadows represent the standard deviation through training with 24 different seeds.}
\end{figure}

At the end of training a model on the source task, the magnitude of back-propagated errors goes toward zero. By switching the task and introducing randomly initialized layers, these errors are suddenly increased. Moreover, the optimization algorithm is usually restarted which causes updates to modify all the parameters with the same rate.
These large back-propagated errors include considerable amounts of noise as shown in Table~\ref{tab:init_noise_regular}. This noise is injected into pre-trained knowledge through the first update.
Fig.~\ref{fig:xl_vars} shows the sudden initial changes in the variance of the input to the last layer when pre-trained models are fine-tuned.

\begin{table}
    \centering
    \caption{Initial percentage of noise energy to total energy of back-propagated error at the output of the last layer, profiled for models fine-tuned using regular fine-tuning; 95\% confidence interval is calculated over 24 seeds.}
    \setlength\tabcolsep{6pt}
    \begin{tabular}{ l|l|l|l|l } 
    & \textbf{MNIST} & \textbf{CIFAR10}  & \textbf{CIFAR100} & \textbf{Caltech101}   \\ \hline
\textbf{ResNet50}~\cite{he2016deep} & $34.60 \pm 0.75$ & $34.66 \pm 0.66$ & $29.64 \pm 0.61$ & $26.11 \pm 1.46$\\
\textbf{ResNet152}~\cite{he2016deep} & $35.16 \pm 0.50$ & $34.81 \pm 0.67$ & $32.25 \pm 0.67$ & $26.15 \pm 1.18$\\
\textbf{DenseNet121}~\cite{Huang_2017} & $33.92 \pm 0.83$ & $35.31 \pm 0.58$ & $29.48 \pm 0.52$ & $27.55 \pm 1.00$\\
\textbf{DenseNet201}~\cite{Huang_2017} & $33.07 \pm 0.81$ & $32.52 \pm 0.81$ & $21.37 \pm 0.74$ & $27.17 \pm 0.80$\\
\textbf{VGG16}~\cite{simonyan2014deep} & $33.39 \pm 1.36$ & $34.98 \pm 0.83$ & $25.18 \pm 0.79$ & $4.97 \pm 1.05$\\
\textbf{VGG19}~\cite{simonyan2014deep} & $32.79 \pm 1.76$ & $34.12 \pm 0.85$ & $23.89 \pm 1.80$ & $4.00 \pm 0.82$\\
\textbf{InceptionV3}~\cite{szegedy2016rethinking} & $32.75 \pm 1.35$ & $33.35 \pm 1.29$ & $24.87 \pm 1.04$ & $23.86 \pm 0.77$\\
    \hline
    \end{tabular}
    \label{tab:init_noise_regular}
\end{table}

The two common approaches to reduce this contamination are to slow down the training and/or to include a warm-up (WU) phase. However, the former slows down the contamination rather than eliminating it~\cite{li2018learning} since the small learning rate also updates the augmented parameters slowly, which injects noise into pre-trained layers for a longer time. In the latter solution, the new parameters are updated for a number of steps before jointly training the entire network. During the warm-up phase, the accuracy of the network is limited since most parts of the network are frozen. Additionally, the effective number of required training steps in the WU phase may be large, depending on the learning rate, initial values of augmented parameters and size of the dataset.

In a more effective approach, we propose an initialization technique for fine-tuning in which the noise is initially trapped only within the task-specific augmented parameters. In contrast to using WU, in our proposed method, the noise is always minimized after the first update and therefore the parameters can be trained altogether afterward. In addition, our method is easier to apply in the sense that we do not manipulate the training process in any way.

\section{Energy Components of Back-propagating Error}

In this section, we first define the energy of the back-propagated error. 
This energy consists of three components from which only one is directly correlated with the accuracy of the estimator and the two others are energies of the true labels and the estimates. We discuss the contribution of these components and find the lower and upper bounds for each one.

Let $\phi_j$, be the sum of energy of $\delta^{L}_j$ through all examples of the batch, defined as
\begin{equation}
\phi_j = \mathop{\mathbb{E}_N} [\left(N{{\delta}^{L}_j}\right)^2] =  \mathop{\mathbb{E}_N}[(\hat{y}_j - y_j)^2].
\label{eq:phi_define}
\end{equation}

Accordingly, 
using Equation~\ref{eq:delta_last}, the total energy of the error over all examples in the batch and all C neurons of the last layer is equal to 
\begin{equation}
    \phi =\sum\limits_{j=1}^{C} \phi_j= \mathop{\mathbb{E}_N} [\hat{\bm{y}}\hat{\bm{y}}^{T}] + \mathop{\mathbb{E}_N} [\bm{y}\bm{y}^{T}] -2  \mathop{\mathbb{E}_N} [\bm{\hat{y}} \bm{y}^{T}].
\end{equation}

Assuming the labels are one-hot encoded; the third term becomes the average probability assignment for the correct labels. The goal of training the model is to maximize this term which is bounded by $0 \leq \mathop{\mathbb{E}_N} [\bm{\hat{y}} \bm{y}^{T}] \leq 1$.
The second term is the energy of the labels and is always equal to one.
Finally, the first term is the energy of the estimates. 
The infimum of this term could be calculated using Cauchy-Schwarz inequality as follows:

\begin{equation}
\left(\bm{\hat{y}}\  \bm{\overrightarrow{1}}    \right)^2 \leq \left(\bm{\overrightarrow{1}}^{T}\  \bm{\overrightarrow{1}} \right)
\left(\hat{\bm{y}}\ \hat{\bm{y}}^{T}\right)\ ,
\end{equation}
\begin{equation}
\bm{\hat{y}}\    \bm{\overrightarrow{1}}   = \sum\limits_{j=1}^{C}{\hat{y}}_{j} = 1 \ ; \ \bm{\overrightarrow{1}}^{T}\  \bm{\overrightarrow{1}} = C \ ,
\end{equation}
\begin{equation}
    \bm{\hat{y}}\   \bm{\hat{y}}^{T} \geq \frac{1}{C} \ .
    \label{eq:sum_estimate_inequality}
\end{equation}
This could also be derived directly from the definition of the softmax (Equation~\ref{eq:sm}), 
\begin{equation}
\zeta = \sum\limits_{j=1}^{C}  {\hat{y}}_j^2 = \sum\limits_{j=1}^{C} \frac{e^{2{a}^{L}_j}}{\left(\sum\limits_{i=1}^{C} e^{{a}^{L}_i}\right)^2} = \frac{ \sum\limits_{j=1}^{C} e^{2{a}^{L}_j}}{\left(\sum\limits_{i=1}^{C} e^{{a}^{L}_i}\right)^2} \ ,
\label{eq:zeta_def}
\end{equation}
followed by taking the partial derivatives with respect to the inputs of the softmax, setting it to zero and re-indexing gives
\begin{equation}
    \frac{\partial \zeta}{\partial {a}^{L}_k} = \frac{2e^{2{a}^{L}_k} \sum\limits_{i=1}^{C} e^{{a}^{L}_i} - 2 e^{{a}^{L}_k} \sum\limits_{j=1}^{C} e^{2\bm{a}^{L}_j} }{\left(\sum\limits_{i=1}^{C} e^{{a}^{L}_i}\right)^3} =0\ ;  
\end{equation}
that results in
\begin{equation}
    \sum\limits_{i=1}^{C} e^{{a}^{L}_i {a}^{L}_k} = \sum\limits_{j=1}^{C} e^{2{a}^{L}_j} \  \ \forall k \ ,
\end{equation}
which means ${a}^{L}_j = {a}^{L}_k\ ; \  j,k \in \{1,2, \ldots, C\}.$
Reforming Equation~\ref{eq:zeta_def} considering equal elements in $\bm{a}^{L}$ finds minimum of $\zeta$ as
\begin{equation}
    \zeta_{min} = \frac{C e^{2{a}^{L}_j}}{\left(e^{{a}^{L}_j}\right)^2} = \frac{1}{C}.
\end{equation}  
The upper bound of the energy of the estimates equals to one and is achieved when their entropy is minimized per example, so $\frac{1}{C}\leq \mathop{\mathbb{E}_N} [\hat{\bm{y}}\hat{\bm{y}}^{T}]\leq 1$.
All in all, the total energy of the back-propagated error is bounded between 0 and 2.

In this section, the bounds for the energy of estimates are investigated. Using the definition of softmax, we showed that to achieve the minimum of $\mathop{\mathbb{E}_N} [\hat{\bm{y}}\hat{\bm{y}}^{T}]$, all the neurons of the last layer should have per-example equal output.
In the next section, we focus on the application of the lower bound found for the energy of the estimates.

\section{ENTAME}
Initially, the energy of the estimates contains pure noise, i.e. it lacks a meaningful relationship with either the inputs or the labels.
We calculated its infimum and showed that it can be achieved when all the estimates are exactly equal to each other for each example.
This condition is intuitively appealing since it maximizes the entropy of the estimates prior to the training when $\bm{y}$ and $\bm{\hat{y}}$ are independent and/or unaligned.
The entropy is exactly reflected in the CE loss which then becomes deterministically $\ln{C}$ regardless of $\bm{\hat{y}}$. 

\begin{figure}[t]  
\centering
\subfloat[Base]{\def\svgwidth{0.38\columnwidth}
\begingroup%
  \makeatletter%
  \providecommand\color[2][]{%
    \errmessage{(Inkscape) Color is used for the text in Inkscape, but the package 'color.sty' is not loaded}%
    \renewcommand\color[2][]{}%
  }%
  \providecommand\transparent[1]{%
    \errmessage{(Inkscape) Transparency is used (non-zero) for the text in Inkscape, but the package 'transparent.sty' is not loaded}%
    \renewcommand\transparent[1]{}%
  }%
  \providecommand\rotatebox[2]{#2}%
  \newcommand*\fsize{\dimexpr\f@size pt\relax}%
  \newcommand*\lineheight[1]{\fontsize{\fsize}{#1\fsize}\selectfont}%
  \ifx\svgwidth\undefined%
    \setlength{\unitlength}{216.28735925bp}%
    \ifx\svgscale\undefined%
      \relax%
    \else%
      \setlength{\unitlength}{\unitlength * \real{\svgscale}}%
    \fi%
  \else%
    \setlength{\unitlength}{\svgwidth}%
  \fi%
  \global\let\svgwidth\undefined%
  \global\let\svgscale\undefined%
  \makeatother%
  \begin{picture}(1,0.54638552)%
    \lineheight{1}%
    \setlength\tabcolsep{0pt}%
    \put(0,0){\includegraphics[width=\unitlength]{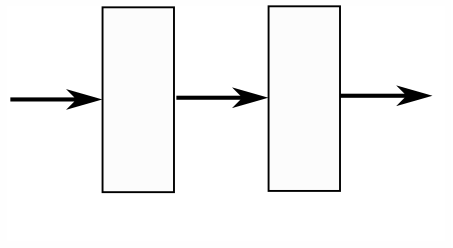}}%
    \put(0.2844704,0.40948433){\color[rgb]{0,0,0}\rotatebox{-90}{\makebox(0,0)[lt]{\lineheight{1.25}\smash{\begin{tabular}[t]{l}FC\end{tabular}}}}}%
    \put(0.01980244,0.39600247){\color[rgb]{0,0,0}\makebox(0,0)[lt]{\lineheight{1.25}\smash{\begin{tabular}[t]{l}$\bm{X}^L$\end{tabular}}}}%
    \put(0.66037373,0.47033193){\color[rgb]{0,0,0}\rotatebox{-90}{\makebox(0,0)[lt]{\lineheight{1.25}\smash{\begin{tabular}[t]{l}softmax\end{tabular}}}}}%
    \put(0.40895318,0.40677536){\color[rgb]{0,0,0}\makebox(0,0)[lt]{\lineheight{1.25}\smash{\begin{tabular}[t]{l}$\bm{A}^L$\end{tabular}}}}%
    \put(0.77317913,0.41810433){\color[rgb]{0,0,0}\makebox(0,0)[lt]{\lineheight{1.25}\smash{\begin{tabular}[t]{l}$\bm{\hat{Y}}$\end{tabular}}}}%
    \put(0.07798884,0.03856658){\color[rgb]{0,0,0}\makebox(0,0)[lt]{\lineheight{1.25}\smash{\begin{tabular}[t]{l}$\bm{W}^{L} \sim \mathcal{N}(0, \  \frac{m}{C})$\end{tabular}}}}%
  \end{picture}%
\endgroup%
\label{fig:regular_arch}}
\subfloat[ENTAME]{\def\svgwidth{0.50\columnwidth}
     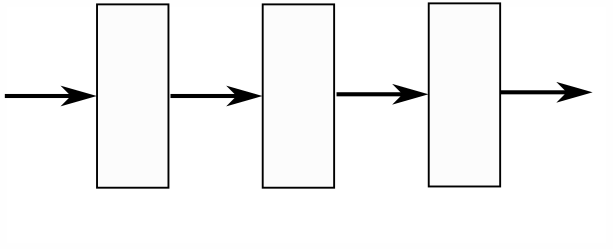\label{fig:entame_arch}}
\caption[architecture]{FNN architecture and last layer's initialization in (a) base model and (b) ENTAME. According to~\cite{he2015delving}, $m$ is 2 for ReLU networks.}
\label{fig:arch}
\end{figure}

Another source of contaminating pre-trained layers is $\bm{W}^L$ itself. This is because $\bm{\delta}^{L-1}$ is affected by both the last layer's error and its weights (see Equation~\ref{eq:delta2delta}). To prevent the noise from contaminating pre-trained layers an efficient solution should consider both of these criteria. We introduce Efficient Neural Task Adaptation by Maximum Entropy initialization (ENTAME) which maximizes the initial entropy of estimates while preventing $\bm{\delta}^{L-1}$ to become contaminated by $\bm{W}^L$. The ENTAME procedure can be described as follows (see Fig.~\ref{fig:arch}).

First, ENTAME requires the features that are fed to the last layer to be normalized. This is done by applying z-normalization across the batch, 
\begin{equation}
    \bm{\bar{x}}^{L}=
    \begin{cases}
      \frac{\bm{x}^{L} - \bm{\mu}_N}{\sqrt{\bm{\sigma}_N^2+\epsilon}}, & \text{if}\ \textit{train} \\\\
      \frac{\bm{x}^{L} - \bm{\hat{\mu}}_N}{\sqrt{\bm{\hat{\sigma}}_N^2+\epsilon}}, & \text{if}\ \textit{inference}
    \end{cases}\ 
    ; \ 
    \bm{\mu}_N, \bm{\hat{\mu}}_N, \bm{\sigma}_N^2, \bm{\hat{\sigma}}_N \in \mathbb{\mathop{R}}^{1\times K} \ ,
    \label{eq:znorm}
\end{equation}

\begin{equation*}
    \bm{\mu}_N = \mathop{\mathbb{E}_N}\left[\bm{x}^{L}\right]\ ; \ \bm{\hat{\mu}}_N = \mathrm{detach}(\mu_N) \ ,
\end{equation*}

\begin{equation*}
    \bm{\sigma}_N^2 = \mathop{\mathbb{E}_N}\left[\left(\bm{x}^{L}_{i} - \bm{\mu}_N\right)^2\right] \ ; \ \bm{\hat{\sigma}}_N = \mathrm{detach}(\bm{\sigma}_N).
\end{equation*}
This is similar to batch-normalization~\cite{ioffe2015batch}, except that it does not need any learnable parameters. The statistics used in inference mode of our simple z-normalization are detached (from the computational graph) version of the ones obtained in the corresponding training step. We discuss the role of
this normalization later in this section.

Second, ENTAME maximizes the entropy of the estimates by initializing the last layer's weights to values drawn from Independent and Identically Distributed (i.i.d.), zero centered normal distribution as follows
\begin{equation}
\bm{W}^{L} \sim \mathcal{N}(0, \  \phi_{w})\ ; \ \phi_{w} = \frac{\gamma^2\lambda^2}{C^2} \ ,
\label{eq:MEI}
\end{equation}
where $\phi_{w}$ is the energy of each element of $\bm{W}^{L}$, $\gamma$ is the initial value of the learning rate (recommended default $\gamma = 10^{-4}$) and $\lambda$ is a hyper-parameter that controls the proportion of noise energy over total energy of last layers weights, right after the first update (recommended range is 1 to 1000). 
$\phi_w$ is chosen to be a numerically small number (for example $\phi_w =10^{-12}$ means that 95\% of the values in $\bm{W}^L$ are initially between $-2\times 10^{-6}$ and $2\times 10^{-6}$), but we will see that such small randomness may help the expressiveness of the model. If the biases are also initialized constantly to all zeros and if $K$ is not extremely large, the energy of $\bm{a}^{L}$ would be initially very small as well. 
Concretely, from the distribution selected for $\bm{W}^L$, the value of each output neuron is approximately zero-centered, or $\mathop{\mathbb{E}_N} \left[\bm{a}^{L}\right] \approx \overrightarrow{\bm{0}}$
and per-example energy of all output neurons together is
\begin{equation}
    \mathop{\mathbb{E}_N} \left[\sum\limits_{j=1}^{C}{a^{L}_j}^2\right] = \mathop{\mathbb{E}_N} \left[ \bm{a}^{L}{\bm{a}^{L}}^{T} \right] =
    \mathop{\mathbb{E}_N} \left[ \bm{\bar{x}} \bm{W}_j^{T} \bm{W}_j \bm{\bar{x}}^{T} \right]= K\phi_w \mathop{\mathbb{E}_N} \left[ \bm{\bar{x}} \bm{\bar{x}}^{T} \right]= \frac{K^2\phi_w}{N}.
    \label{eq:a_ij_0}
\end{equation}
This is derived from $\mathop{\mathbb{E}_N}[\bm{\bar{x}} \bm{\bar{x}}^{T}] = \frac{K}{N}$, due to $\bm{\bar{X}}$ being normalized across the batch.

Moreover, the exponential function is close to linear when its input is close to zero. This could be easily shown by using the result of Equation~\ref{eq:a_ij_0}, and the Taylor series approximation of the exponential function around zero, $e^{{a}^{L}_j} \approx 1+ {a}^{L}_j$.
Plugging this into the softmax definition yields 
${\hat{y}_j} \approx \frac{1}{C}$,
which approximately maximizes the entropy as desired. 

When the estimates are equal for each example, $\delta_j^L$ becomes only a function of the prior, i.e.,
\begin{equation}
    \delta^{L}_{j}=
    \begin{cases}
      \frac{1}{NC}, & \text{if}\ y_{j}=0 \\\\
      \frac{1}{N}\left(\frac{1}{C}-1\right), & \text{if}\ y_{j}=1 
    \end{cases}
    \ \ ; \ \  j \in \{1, 2, \ldots, C\}.
    \label{eq:delta_me_cases}
\end{equation}
Accordingly, the gradients of the loss with respect to last layer's parameters are simplified to
\begin{equation}
    \nabla_{\bm{W}^{L}_j} \Lagr = \mathop{\mathbb{E}_N} \left[(\frac{1}{C}- {y_j}) \ \bm{\bar{x}}^{L}\right]\ ; \ \nabla_{{b}^{L}_j} \Lagr = \mathop{\mathbb{E}_N} \left[\frac{1}{C}- {y_j}\right].
    \label{eq:updates_L_norm}
\end{equation}
Applying the updates results in
\begin{equation}
    \bm{W}^{L,1}_j = \bm{W}^{L,0}_j + \gamma \mathop{\mathbb{E}_N} \left[{y_j} \ \bm{\bar{x}}^{L}\right] - \gamma \frac{1}{C} \mathop{\mathbb{E}_N} \left[ \bm{\bar{x}}^{L}\right]\ ; {b}^{L,1}_j = \gamma \mathop{\mathbb{E}_N} \left[{y_j}\right] - \gamma \frac{1}{C} \ ,
    \label{eq:updated_params}
\end{equation}
where $\gamma$ is the initial learning rate and the second number in the superscripts represent the number of updates, e.g. $\bm{W}^{l,u}$ indicates to the weights of the $l$-th layer after $u$ updates.
The error can not move further backward at this point since $\phi_w$ is very small, making $\bm{\delta}^{L-1}$ negligible. 

After the first update, outputs of each neuron of the last layer can get comparably high expected values. 
This may cause the estimates to have much lower entropy compared to the initial state. On the other hand, depending on $\bm{Y}$ and $\bar{\bm{X}}$, multiple rows and columns of weights and corresponding elements of biases from the last layer can possibly get identical first updates.
This may cause the entropy of the estimates to stay comparably high for each example. Very small random numbers, used to initialize $\bm{W}^L$, help these identical estimates to diverge and make different estimates. Notice that as the expectation of ${a}^L_j$ gets away from zero toward positive values, the exponential functions in softmax make the small difference much larger. Therefore, the expressiveness of the last layer is preserved by initializing its weights to very small numbers instead of zero.  

The first update makes the energy of $\bm{W}^L$ large enough to let the error of the next updates back-propagate through it and reach the pre-trained layers. In other words, this automatically opens up the stalled way and lets the error to back-propagate to the output of other layers. This is enough for correctly guiding pre-trained parameters with an advanced optimization algorithm like Adam~\cite{kingma2015adam} is used. Most of the noise is purified and the next back-propagated errors toward pre-trained features are meaningful and contain both prior and likelihood. In more details, the energy of the $j$-th row in $\bm{W}^{L,1}$ becomes

\begin{equation}
    \bm{W}^{L,1}_j {\bm{W}^{L,1}_j}^{T} = K\phi_w + \gamma^2 \mathop{\mathbb{E}_N} \left[\left( \frac{1}{C} - y_j \right)^2 \bar{\bm{x}} \  \bar{\bm{x}}^{T}\right]
\end{equation}
\begin{equation*}
    = K \gamma^2 \frac{\lambda^2}{C^2} + K \gamma^2 \frac{1}{NC^2} + \gamma^2 \frac{C-2}{NC} \sum\limits_{i=1}^{N} \bm{Y}_i \  \bar{\bm{X}_i} \  \bar{\bm{X}_i}^{T} \ ,
\end{equation*}
which means
\begin{equation}
    \frac{K \gamma^2}{C^2} \left(\lambda^2
    + \frac{1}{N}\right) \leq \bm{W}^{L,1}_j {\bm{W}^{L,1}_j}^{T} \leq \frac{K \gamma^2}{C^2} \left(\lambda^2
    + \frac{1}{N} + \frac{C(C-2)}{N}\right).
\end{equation}
Energy of $\bm{W}^{L,1}_j$ is $\frac{K}{N}$ times the energy of $j$-th bias in the last layer. Since ${b}^{L,1}_j$ contains only information about the prior, we desire to make its energy initially smaller than $\bm{W}^{L,1}_j$. For this to happen, we have to choose initial $N$ such that $N<K$ (which usually is satisfied).
On the other hand, $\lambda$ should be chosen such that $\lambda^2 < \frac{1}{N}$ but not very small to numerically reduce the rank of $\bm{W}^{L,1}_j$ due to possible similar updates (see Equation~\ref{eq:updated_params}), which may results in higher entropy of $\bm{\hat{y}}$.
$N\lambda^2$ roughly determines the maximum proportion of remaining energy of noise to the total energy of elements of $\bm{W}^{L,1}$.

\subsection{The role of feature normalization}
Applying z-normalization on top of features may increase or decrease the level of average feature-wise energy in $\bm{X}^{L}$ resulting in less need for tweaking the learning rate and $\phi_w$ for different tasks and even different models. If the values of $\bm{X}^{L}$ are too small, it may take a longer time for $\bm{W}^{L}$ to grow which leaves pre-trained features unchanged for a longer time. 

Z-normalization across batches plays a more important role than just equalization. 
To clarify, lets assume that we are going to do image classification and two images in a batch contain exactly the same pattern or visual object. If one of the columns of $\bm{X}^{L}$ represents a feature that recognizes said pattern, we expect the feature to reflect the presence of the pattern in both of the mentioned images equally. The problem is that raw inputs are usually normalized with statistics that are identically applied on all pixels in all examples. In the best case, such normalization is applied separately for different channels. Object-wise normalization does not seem to be feasible prior to detection which indirectly is done through training neural network classifier. Therefore, even if the same object is exactly copied in both images due to normalizing raw images, one object may get less intensified than the other one. This may directly be reflected to the values of the particular column of $\bm{X}^L$ responsible for showing the presence of the desired object. Z-normalization, compensates for this problem by normalizing the features after they are detected.

Batch-normalization layers used in between the hidden layers of some pre-trained models, usually need more training steps to adapt  to the distribution of target task's data. Since we  also care about the performance of the model in the first training steps, feeding normalized features to the last layer is vital.
Our simple z-normalization applied on $\bm{X}^L$,
directly influences the first update of $\bm{W}^L$.

\section{Experiments on Natural Images}
ImageNet~\cite{russakovsky2015imagenet} ILSVRC 2012 is the source dataset used to pre-train the models. We fine-tune each pre-trained model on the following datasets: MNIST~\cite{lecun1998gradient}, CIFAR10, CIFAR100~\cite{krizhevsky2009learning} and Caltech101~\cite{fei2007learning}. 
The latter dataset is not originally separated into train and test nor is balanced in contrast to the other ones. We randomly split each Caltech101 category into train and test subsets with 15 percent chance of drawing each image for test subset. 

Prior to feeding the input to the models, each channel is normalized with its mean and standard deviation obtained from all pixels of that channel throughout the corresponding training subset. Training images are also augmented with random horizontal flip. 

The set of used architectures are listed in the leftmost column of Table~\ref{tab:acc}. Among these models InceptionV3 requires all images to be scaled up to $229\times 229$, so due to limitations in device's memory the batch size of 64 is chosen for this architecture. In addition, the other models that are trained on the Caltech101 dataset are also fed 64 images per batch owning to large image sizes. All other models and dataset use batch size of 256.

The initialization recommended by~\cite{he2015delving} is used for augmented layers in the base models. 
We tried to unify the problem by applying similar conditions for training different models as much as possible. This by itself would show the impact of the proposed method and how universally it could help the task adaptation, even without considering hyper-parameter tuning. Accordingly, learning rate is set to 0.0001 for all models and datasets and the value of $\phi_w$ is chosen to be $10^{-12}$ everywhere. 

Table~\ref{tab:acc} shows the progress of test accuracy of pre-trained models fine-tuned on each dataset. The smaller plot inside each larger one shows the same curves zoomed-in the first steps of training. The colorful shade around each curve shows the standard deviation across 24 different seeds. Each plot includes 4 curves color mapped as follows; blue: base, orange: base with a single Warm Up (WU) step, green: ENTAME's Maximum Entropy Initialization (MEI), red: full ENTAME or MEI + Feature Normalization (FN). We also did  experiments with only applying FN, but they mostly perform worse than all other cases, so we don't include them to save space and make plots more readable.

\begin{table}
\centering
 \caption{Test accuracy progress for fine-tuning models that are pre-trained on ImageNet dataset. The horizontal axes on each plot show the number of training steps. Colorful shades show the standard deviation across different seeds. A superscript $*$ means that all models in corresponding row or column are trained with batch size of 64 instead of 256 to make the model fit into the device. The smaller plots inside the bigger ones are just zoomed-in version of the same curves for the first few steps.}
\begin{tabular}{lccc}
\toprule
\includegraphics[scale=0.06,valign=c]{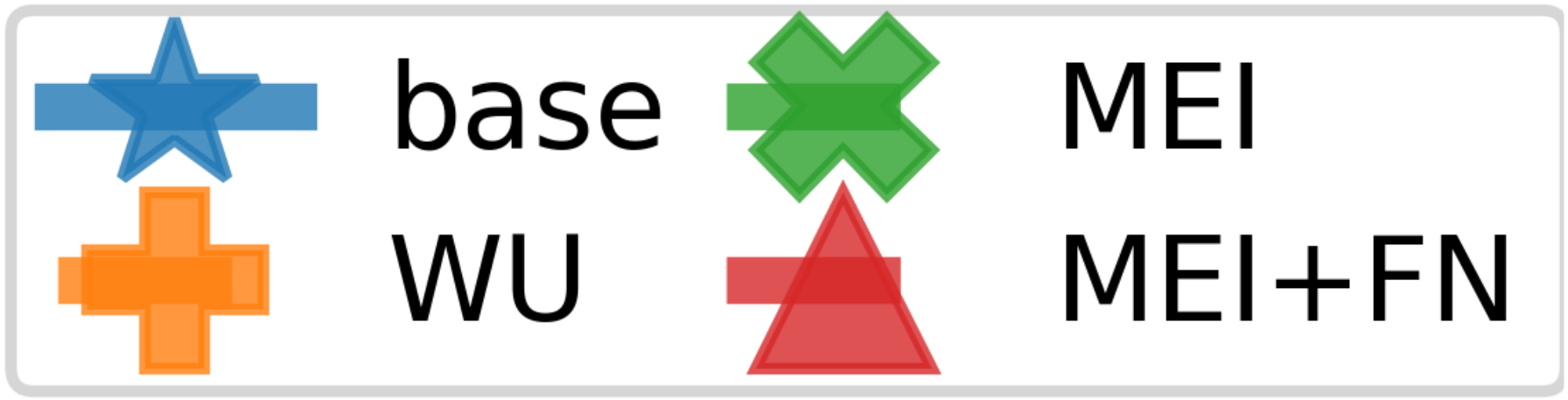} & \textbf{CIFAR10}  & \textbf{CIFAR100} & \textbf{{Caltech101}$^*$} \\
\midrule
\textbf{ResNet50}~\cite{he2016deep} &
\includegraphics[scale=0.14,valign=c]{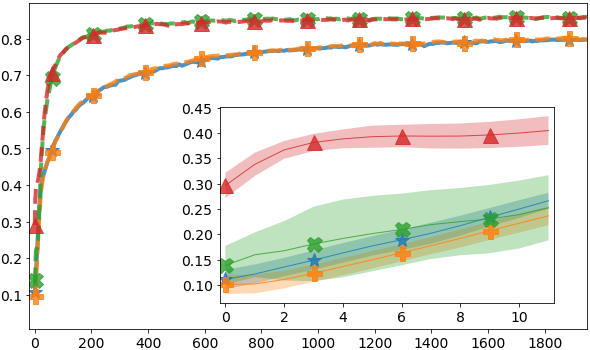} &
\includegraphics[scale=0.14,valign=c]{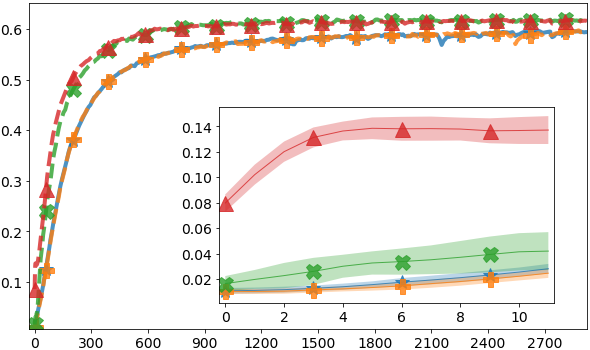} &
\includegraphics[scale=0.14,valign=c]{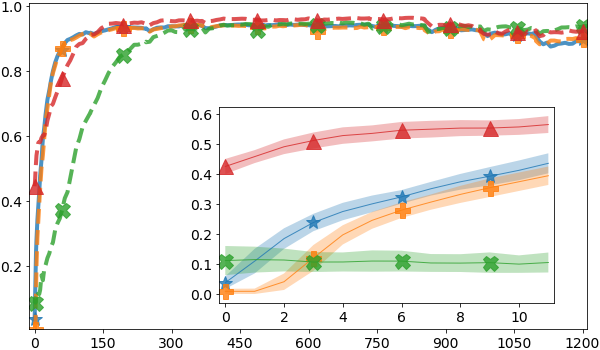}
\\
\textbf{ResNet152}~\cite{he2016deep} &
\includegraphics[scale=0.14,valign=c]{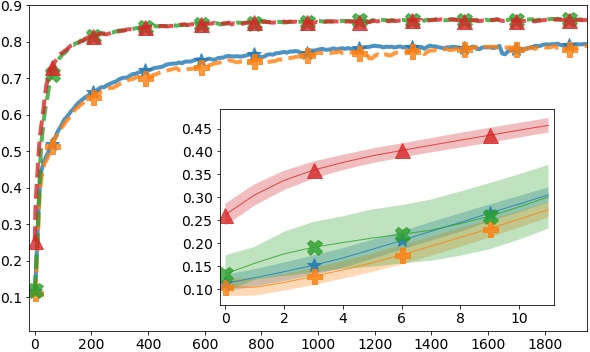} &
\includegraphics[scale=0.14,valign=c]{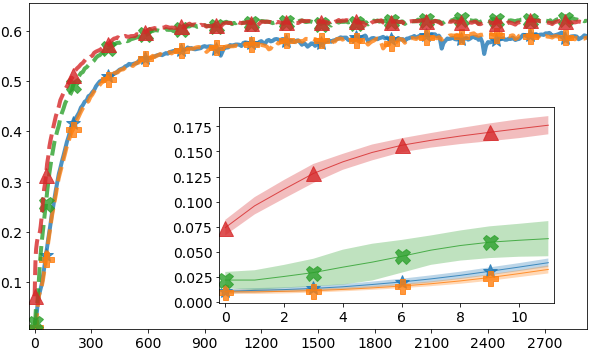} &
\includegraphics[scale=0.14,valign=c]{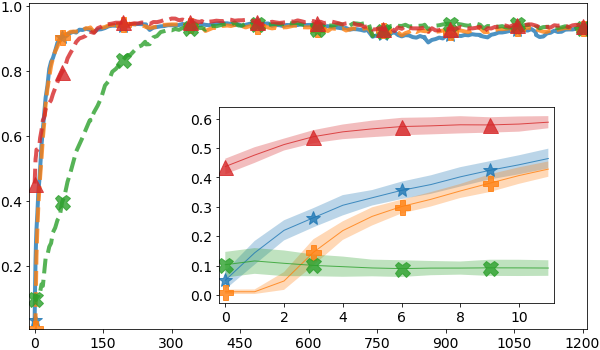}
\\
\textbf{DenseNet121}~\cite{Huang_2017} &
\includegraphics[scale=0.14,valign=c]{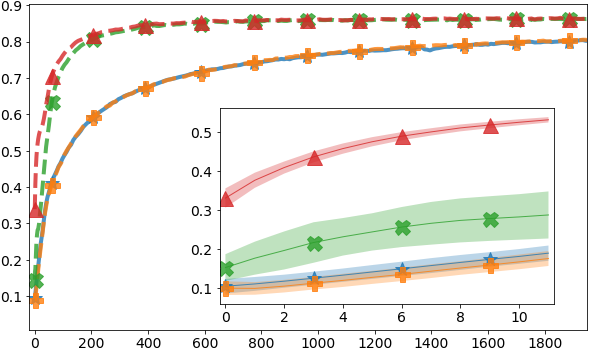} &
\includegraphics[scale=0.14,valign=c]{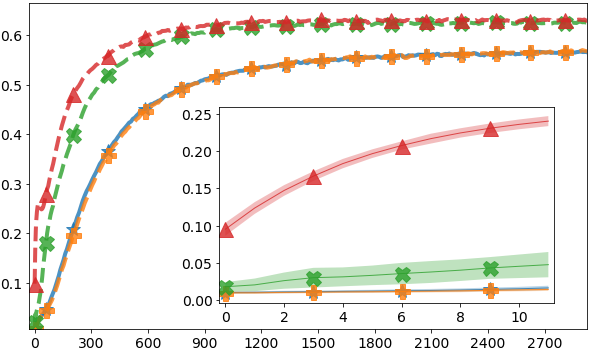} &
\includegraphics[scale=0.14,valign=c]{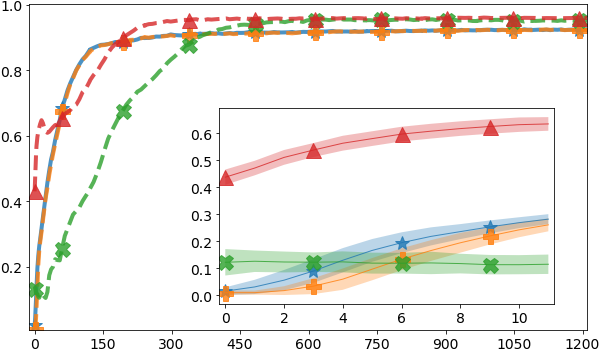}
\\
\textbf{DenseNet201}~\cite{Huang_2017} &
\includegraphics[scale=0.14,valign=c]{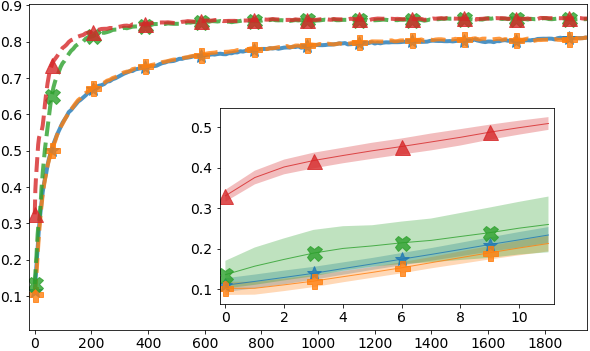} &
\includegraphics[scale=0.14,valign=c]{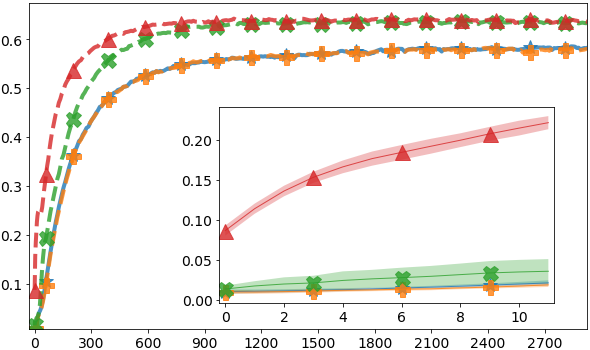} &
\includegraphics[scale=0.14,valign=c]{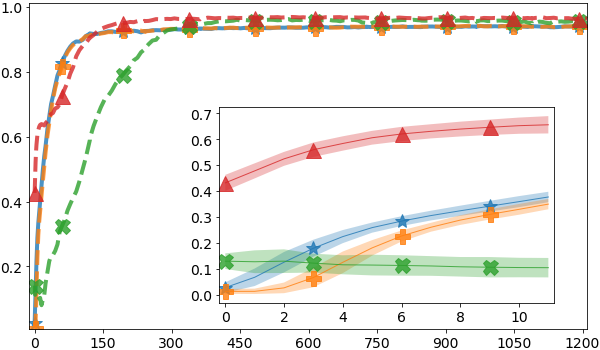}
\\
\textbf{VGG16}~\cite{simonyan2014deep} &
\includegraphics[scale=0.14,valign=c]{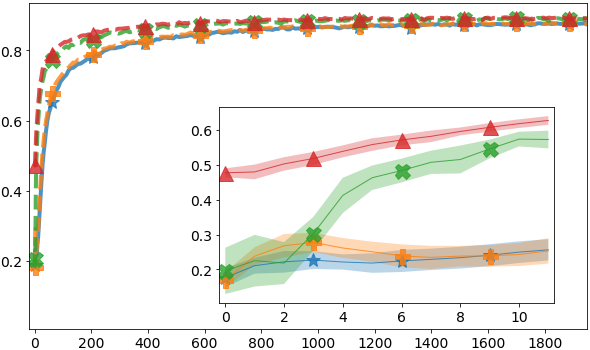} &
\includegraphics[scale=0.14,valign=c]{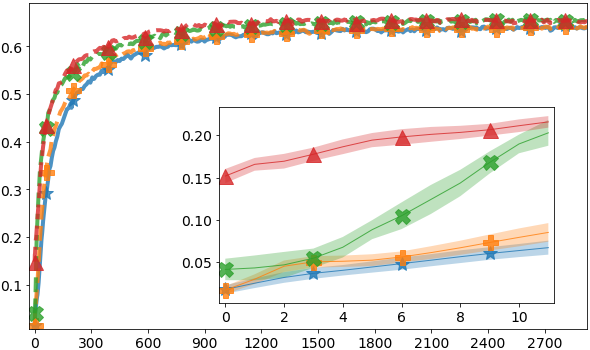} &
\includegraphics[scale=0.14,valign=c]{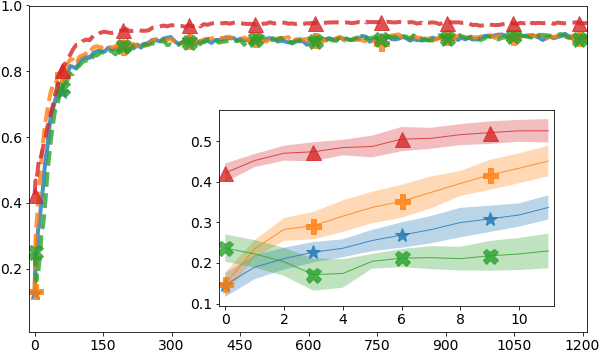}
\\
\textbf{VGG19}~\cite{simonyan2014deep} &
\includegraphics[scale=0.14,valign=c]{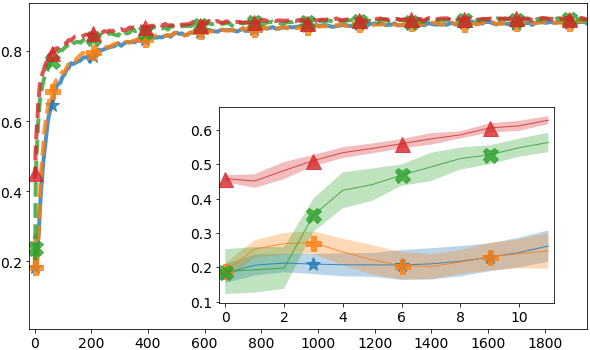} &
\includegraphics[scale=0.14,valign=c]{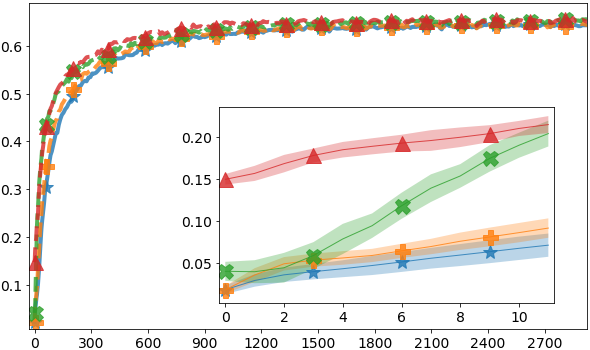} &
\includegraphics[scale=0.14,valign=c]{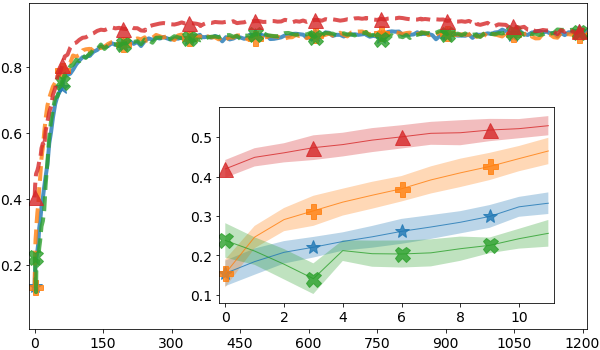}
\\
\textbf{InceptionV3$^*$}~\cite{szegedy2016rethinking} &
\includegraphics[scale=0.14,valign=c]{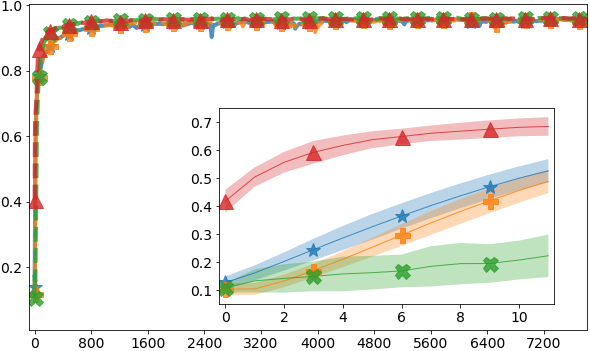} &
\includegraphics[scale=0.14,valign=c]{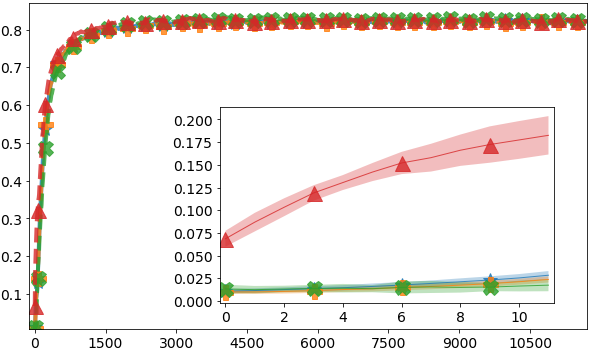} &
\includegraphics[scale=0.14,valign=c]{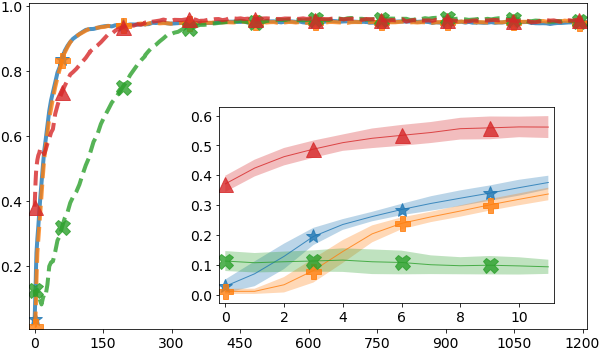}
\\
 \end{tabular}
 \label{tab:acc}
 \end{table}

To measure how the convergence is sped up initially, we compare average progressing accuracy over first ten training steps. Paired t-test suggests that ENTAME significantly enhances the test accuracy compared to the base method for all architectures and datasets mentioned in this paper. Table~\ref{tab:speedup} shows the average increase in the accuracy of the first 10 training steps with 95\% confidence. We have observed further improvements by adjusting 
$\lambda$ and the batch size but to show the robustness of our model we tried to keep the same setup as much as possible. 

\begin{table}
    \centering
    \caption{Average initial test accuracy improvement by using ENTAME instead of base method. The entries show increase in the mean of test accuracy over first 10 steps of training with 95\% confidence calculated over 24 seeds.}
    \setlength\tabcolsep{6pt}
    \begin{tabular}{ l|l|l|l|l } 
    & \textbf{MNIST} & \textbf{CIFAR10}  & \textbf{CIFAR100} & \textbf{Caltech101}   \\ \hline
    \textbf{ResNet50}\cite{he2016deep} & $10.86 \pm 2.97$ &  $21.81 \pm 1.10$ &  $10.19 \pm 0.32$ &  $31.29 \pm 1.21$\\
    \textbf{ResNet152}\cite{he2016deep} & $4.52 \pm 1.90$ &  $18.94 \pm 1.22$ &  $9.74 \pm 0.41$ &  $30.75 \pm 1.09$\\
    \textbf{DenseNet121}\cite{Huang_2017} & $12.61 \pm 1.55$ &  $28.38 \pm 0.98$ &  $13.21 \pm 0.26$ &  $43.95 \pm 1.25$\\
    \textbf{DenseNet201}\cite{Huang_2017} & $17.90 \pm 1.85$ &  $26.10 \pm 1.08$ &  $11.99 \pm 0.26$ &  $39.09 \pm 1.95$\\
    \textbf{VGG16}\cite{simonyan2014deep} & $35.29 \pm 2.40$ &  $29.14 \pm 0.78$ &  $13.95 \pm 0.38$ &  $25.86 \pm 0.90$\\
    \textbf{VGG19}\cite{simonyan2014deep} & $33.04 \pm 1.69$ &  $28.37 \pm 1.30$ &  $13.38 \pm 0.40$ &  $25.58 \pm 1.31$\\
    \textbf{InceptionV3}\cite{szegedy2016rethinking} & $9.17 \pm 1.38$ &  $33.21 \pm 2.02$ &  $8.90 \pm 0.38$ &  $31.94 \pm 1.48$\\

    \hline
    \end{tabular}
    \label{tab:speedup}
\end{table}

Finally the converged accuracy of each curve shown in Table~\ref{tab:acc} is listed in Tables 
\ref{tab:cifar10_acc}, \ref{tab:cifar100_acc} and \ref{tab:caltech101_acc}. The convergence test accuracy is recorded after training models for 10 epochs if target dataset is CIFAR10 or Caltech101 and 15 epochs if target dataset is CIFAR100.
We did further experiments on ResNet\cite{he2016deep}, DenseNet\cite{Huang_2017} and VGG\cite{simonyan2014deep} with other popular sizes but similar results were obtained so we only reported results of the two most common sizes of each in the above-mentioned tables.

\begin{table}[t]
    \centering
    \caption{Convergence test accuracy of models trained on CIFAR10 dataset with 95\% confidence.}
    \setlength\tabcolsep{6pt}
    \begin{tabular}{ l|c|c||c|c }
    & \textbf{Base} & \textbf{Base+WU}  & \textbf{MEI} & \textbf{MEI+FN}   \\ \hline
    \textbf{ResNet50}\cite{he2016deep} & $79.73 \pm 0.43$ & $79.74 \pm 0.76$ & $ \bm{85.80 \pm 0.35}$ & $\bm{85.54 \pm 0.20}$\\
    \textbf{ResNet152}\cite{he2016deep} & $79.18 \pm 1.07$ & $78.70 \pm 1.16$ & $\bm{86.02 \pm 0.32}$ & $\bm{86.01 \pm 0.14}$\\
    \textbf{DenseNet121}\cite{Huang_2017} & $80.21 \pm 0.23$ & $80.39 \pm 0.31$ & $\bm{86.20 \pm 0.23}$ & $\bm{86.32 \pm 0.27}$\\
    \textbf{DenseNet201}\cite{Huang_2017} & $81.11 \pm 0.35$ & $80.92 \pm 0.46$ & $\bm{86.27 \pm 0.20}$ & $\bm{86.38 \pm 0.29}$\\
    \textbf{VGG16}\cite{simonyan2014deep} & $87.50 \pm 0.37$ & $87.70 \pm 0.47$ & $\bm{88.79 \pm 0.61}$ & $\bm{89.19 \pm 0.25}$\\
    \textbf{VGG19}\cite{simonyan2014deep} & $87.94 \pm 0.60$ & $88.12 \pm 0.25$ & $\bm{88.77 \pm 0.22}$ & $\bm{89.12 \pm 0.19}$\\
    \textbf{InceptionV3}\cite{szegedy2016rethinking} & $95.58 \pm 0.47$ & $95.50 \pm 0.60$ & $95.93 \pm 0.27$ & $95.91 \pm 0.21$\\    \hline
    \end{tabular}
    \label{tab:cifar10_acc}
\end{table}

\begin{table}[t]
    \centering
    \caption{Convergence test accuracy of models trained on CIFAR100 dataset with 95\% confidence.}
    \setlength\tabcolsep{6pt}
    \begin{tabular}{ l|c|c||c|c }
    & \textbf{Base} & \textbf{Base+WU}  & \textbf{MEI} & \textbf{MEI+FN}   \\ \hline 
    \textbf{ResNet50}\cite{he2016deep} & $59.38 \pm 0.39$ & $59.44 \pm 0.41$ & $\bm{61.50 \pm 0.46}$ & $\bm{61.54 \pm 0.35}$\\
    \textbf{ResNet152}\cite{he2016deep} & $58.71 \pm 1.37$ & $58.50 \pm 0.89$ & $\bm{61.91 \pm 0.63}$ & $\bm{61.85 \pm 0.77}$\\
    \textbf{DenseNet121}\cite{Huang_2017} & $56.52 \pm 0.46$ & $56.70 \pm 0.26$ & $\bm{62.52 \pm 0.41}$ & $\bm{62.90 \pm 0.27}$\\
    \textbf{DenseNet201}\cite{Huang_2017} & $58.27 \pm 0.62$ & $57.98 \pm 0.58$ & $\bm{63.36 \pm 0.15}$ & $\bm{63.64 \pm 0.59}$\\
    \textbf{VGG16}\cite{simonyan2014deep} & $63.94 \pm 0.24$ & $63.77 \pm 0.22$ & $\bm{65.19 \pm 0.58}$ & $\bm{64.99 \pm 0.40}$\\
    \textbf{VGG19}\cite{simonyan2014deep} & $64.30 \pm 0.42$ & $64.47 \pm 0.54$ & $65.11 \pm 0.28$ & $65.02 \pm 0.21$\\
    \textbf{InceptionV3}\cite{szegedy2016rethinking} & $82.25 \pm 0.37$ & $82.19 \pm 0.21$ & $82.17 \pm 0.32$ & $81.75 \pm 0.64$\\
    \hline
    \end{tabular}
    \label{tab:cifar100_acc}
\end{table}

\begin{table}[t]
    \centering
    \caption{Convergence test accuracy of models trained on Caltech101 dataset with 95\% confidence.}
    \setlength\tabcolsep{6pt}
    \begin{tabular}{ l|c|c||c|c }
    & \textbf{Base} & \textbf{Base+WU}  & \textbf{MEI} & \textbf{MEI+FN}   \\ \hline 
    \textbf{ResNet50}~\cite{he2016deep} & $89.69 \pm 3.30$ & $90.69 \pm 1.12$ & $\bm{93.87 \pm 0.61}$ & $92.15 \pm 1.61$\\
    \textbf{ResNet152}~\cite{he2016deep} & $93.12 \pm 1.01$ & $92.98 \pm 1.04$ & $93.19 \pm 0.84$ & $93.71 \pm 1.34$\\
    \textbf{DenseNet121}~\cite{Huang_2017} & $92.36 \pm 0.67$ & $92.41 \pm 0.91$ & $\bm{95.13 \pm 1.03}$ & $\bm{95.96 \pm 0.19}$\\
    \textbf{DenseNet201}~\cite{Huang_2017} & $93.95 \pm 0.53$ & $94.01 \pm 0.50$ & $94.87 \pm 1.90$ & $\bm{96.50 \pm 0.71}$\\
    \textbf{VGG16}~\cite{simonyan2014deep} & $89.35 \pm 1.75$ & $91.02 \pm 0.95$ & $90.07 \pm 0.68$ & $\bm{94.69 \pm 1.19}$\\
    \textbf{VGG19}~\cite{simonyan2014deep} & $90.50 \pm 1.42$ & $89.96 \pm 0.90$ & $89.70 \pm 1.28$ & $90.53 \pm 4.15$\\
    \textbf{InceptionV3}~\cite{szegedy2016rethinking} & $94.98 \pm 0.54$ & $95.07 \pm 0.67$ & $95.70 \pm 0.41$ & $95.50 \pm 0.89$\\
    \hline
    \end{tabular}
    \label{tab:caltech101_acc}
\end{table}


\end{document}